\documentclass{article}
\usepackage[T1]{fontenc}
\usepackage{spconf,amsmath,graphicx,hyperref,booktabs,multirow,multicol}

\makeatletter
\let\oldthebibliography\thebibliography
\def\thebibliography#1{%
  \oldthebibliography{#1}%
  \fontsize{9pt}{10pt}\selectfont    
  \setlength{\itemsep}{0pt plus 0.2pt}%
  \setlength{\parskip}{0pt}%
  \setlength{\parsep}{0pt}%
}
\makeatother

\title{Assessing Identity Leakage in Talking Face Generation: Metrics and Evaluation Framework}
\name{\parbox{\linewidth}{\centering Dogucan Yaman\textsuperscript{1,2} \qquad Fevziye Irem Eyiokur\textsuperscript{1,2} \qquad Hazım Kemal Ekenel\textsuperscript{3} \qquad Alexander Waibel\textsuperscript{1,2,4} }}

\address{\textsuperscript{1} Karlsruhe Institute of Technology, \textsuperscript{2} KIT Campus Transfer GmbH (KCT), \\ \textsuperscript{3} Istanbul Technical University, \textsuperscript{4} Carnegie Mellon University  }

\begin{document}
%
\maketitle
\begin{abstract}
Video editing-based talking face generation aims to preserve video details such as pose, lighting, and gestures while modifying only lip motion, often using an identity reference image to maintain speaker consistency. 
However, this mechanism can introduce lip leakage, where generated lips are influenced by the reference image rather than solely by the driving audio. 
Such leakage is difficult to detect with standard metrics and conventional test setup. 
To address this, we propose a systematic evaluation methodology to analyze and quantify lip leakage. 
Our framework employs three complementary test setups: silent-input generation, mismatched audio–video pairing, and matched audio–video synthesis.
We also introduce derived metrics including lip-sync discrepancy and silent-audio-based lip-sync scores. 
In addition, we study how different identity reference selections affect leakage, providing insights into reference design.
The proposed methodology is model-agnostic and establishes a more reliable benchmark for future research in talking face generation.
\end{abstract}
\section{Introduction}
\label{sec:intro}
Audio-driven video editing-based talking face generation~\cite{prajwal2020lip} aims to preserve the overall video and facial details (e.g., pose, lighting, gestures) while modifying only the lip region to match target speech. 
This property makes inpainting particularly attractive for applications such as movie dubbing, where strict preservation of non-lip details is required and cannot be achieved with one-shot generation or portrait matting.
Numerous works have focused on improving visual quality~\cite{yamamoto1998lip,cheng2022videoretalking,mukhopadhyay2024diff2lip,zhang2023dinet,stypulkowski2024diffused,yaman2025mask,wang2023lipformer}, identity preservation~\cite{zhong2023identity,yaman2025mask}, and lip-sync accuracy~\cite{yaman2024audio,yaman2024audiodriventalkingfacegeneration,muaz2023sidgan,park2022synctalkface,wang2023seeing}. While some approaches integrate talking face generation into end-to-end video translation systems~\cite{ritter1999face,waibel2023face,eyiokur2024titanic}, which are part of the drive to go from speech translation to language transparency~\cite{waibel2016hybrid,niehues2018low}, others adopt one-shot talking head generation for avatar creation~\cite{zhou2020makelttalk,song2022everybody,zhang2021flow,zhang2023sadtalker,zhang2024emodiffhead,zhang2024emotalker,tian2024emo}.
Standard approaches operate on a frame-by-frame basis, masking the lower face during training to hide ground-truth (GT) lip motion. 
Since the model lacks information about the masked region, an identity reference image is provided to maintain speaker identity. 
Typically, this reference is randomly chosen from another frame in the same video, ensuring consistent appearance but different lip motion. 
At inference time, some models select the current input frame as the reference~\cite{prajwal2020lip,yaman2024audio,yaman2024audiodriventalkingfacegeneration,waibel2023face}, while others rely on a randomly chosen frame or the first frame~\cite{zhong2023identity}.

While effective for identity preservation, the use of reference images introduces a subtle yet important vulnerability: generated lip motion can be influenced by the reference itself, rather than being determined solely by the input audio. 
This phenomenon is referred to as \textit{lip leakage} in the literature~\cite{yaman2024audiodriventalkingfacegeneration}.
To mitigate it, some methods employ multiple reference images~\cite{zhong2023identity}, enhancing feature robustness through greater variation and reducing leakage from any single reference.
Moreover, some methods propose to modify the identity reference image (silent face or canonical face) to mitigate lip leakage~\cite{yaman2024audiodriventalkingfacegeneration,cheng2022videoretalking}.
In \cite{bigata2025keysync}, it is proposed to measure mouth aperture ratio as a leakage metric by using silent audio and non-silent video.
However, existing test setups and evaluation metrics such as lip-sync accuracy and visual quality are insufficient to detect such leakage, as a system may achieve high scores even when the generation process is biased by reference lip motion.
Similarly, when models exhibit poor or suboptimal performance, the current evaluation pipeline makes it difficult to determine whether the cause is lip leakage.
This leakage also poses practical risks in real applications where controllability and reliability are essential, including virtual avatars, human–computer interaction, and video \& movie dubbing. 
If lip movements are unintentionally guided by the identity reference image, the output may appear visually synchronized but semantically misaligned with the audio if the most common matched audio-video test setup is employed, undermining both the validity of scientific benchmarks and the trustworthiness of deployed systems.

To address this gap in the field, we propose a systematic evaluation methodology for detecting lip leakage from identity reference.
Our framework consists of three complementary test setups: silent-input generation, mismatched audio–video pairing, and matched audio–video synthesis.
These test setups reveal hidden leakage behaviors. 
We further introduce derived metrics, including (i) lip-sync discrepancy (LSD) between matching and non-matching conditions, and (ii) lip-sync scores computed from silent-input generations compared against original audio. 
Finally, we analyze how different choices of identity reference frames (first frame vs. current frame) affect leakage, providing insights into reference selection criteria. 
Our methodology is model-agnostic, easy to implement, and provides a more reliable benchmark for advancing talking face generation research.
Our contributions are as follows:
(1) We propose a test methodology to analyze lip leakage from identity references, evaluate model robustness to reference selection, and examine the impact of reference image choice.
(2) We introduce three metrics within the proposed test methodology to quantitatively assess lip leakage.
(3) Our test methodology evaluates not only lip leakage but also visual sensitivity.
(4) We conduct extensive experiments and evaluate the models using benchmark metrics to assess their capabilities.



\begin{table*}[]
    \centering
    \resizebox{\linewidth}{!}{\begin{tabular}{c|l|c|l}
    \toprule
    Setup & Name & Abbreviation & Description \\
    \midrule
    \multirow{4}{*}{Metrics} & Silent LSE-C & $LSE-C_S$ & LSE-C applied on silent-input generations, compared against original audio. \\
    & Silent LSE-D & $LSE-D_S$ & LSE-D applied on silent-input generation, compared against original audio. \\
    & Lip-Sync Discrepancy w/ Current Ref. & LSD-CR & Difference in lip-sync scores between audio-matched (AM) and audio-mismatched (XM) setups, with current-frame reference. \\
    & Lip-Sync Discrepancy w/ Alternative Ref. & LSD-AR & Same as LSD-CR, but with alternative reference (first frame / random / multi-frame). \\
    \midrule
    \multirow{3}{*}{Generation Setup} & Audio-Matched & AM & Video generated with its corresponding audio. \\
    & Audio-Mismatched & XM & Video generated with randomly paired, non-matching audio. \\
    & Silent-Input Generation & SI & Video generated using silent audio as input, used to probe lip leakage. \\
    \midrule
    \multirow{2}{*}{Reference Selection} & Current Reference & CR & Identity reference is the same as the masked input frame. \\
    & Alternative Reference & AR & Identity reference chosen from first frame or other strategies if proposed by the authors (e.g., random frame, multiple random frame). \\
    \bottomrule
    \end{tabular}}
    \caption{Summary of proposed evaluation metrics and setups.}
    \label{tab:metrics}
\end{table*}

\begin{figure}
    \centering
    \includegraphics[width=0.8\linewidth]{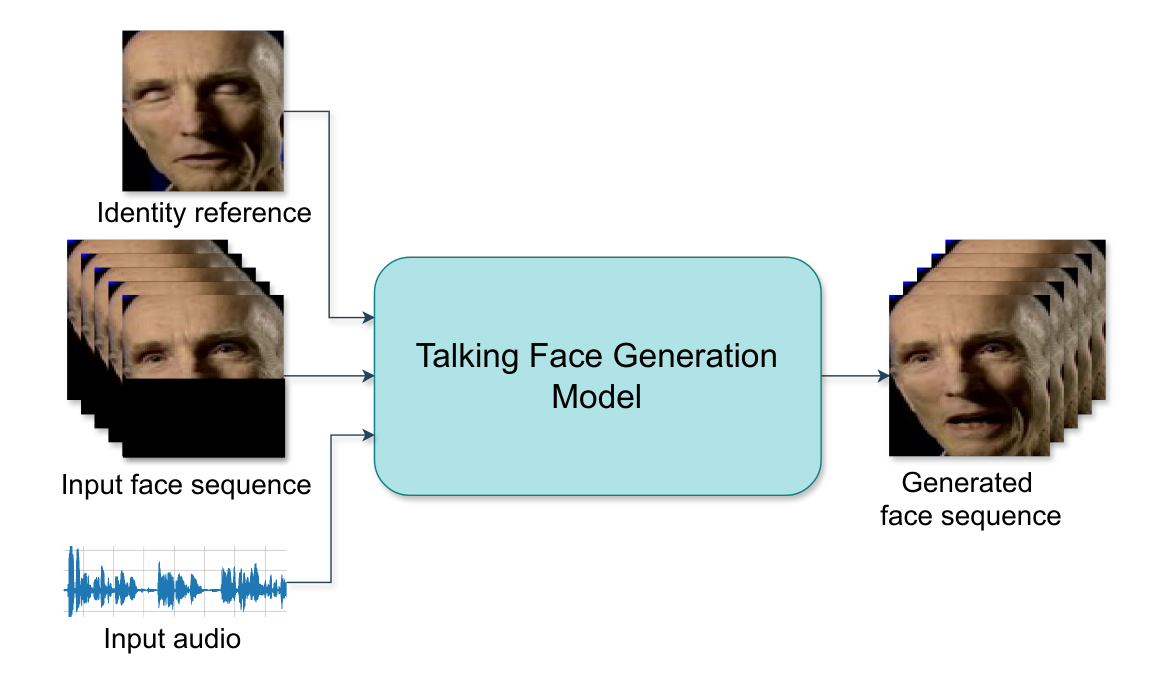}
    \caption{Standard pipeline for audio-driven video editing-based talking face generation.}
    \label{fig:TFG_pipeline}
\end{figure}

\section{Methodology}
\label{sec:method}

We propose a systematic evaluation framework to analyze lip leakage from identity references in talking face generation. 
Our methodology comprises three main components: audio–video generation setups, identity reference selection strategies, and lip leakage assessment metrics. 
Together, these provide a comprehensive, model-agnostic framework to quantify lip leakage and evaluate model robustness in preserving identity and visual quality. 
By systematically varying audio and reference conditions, our framework uncovers subtle leakage behaviors that standard lip-sync and visual quality metrics may miss. 
Table \ref{tab:metrics} summarizes the proposed metrics, methodology, and setups, while Figure \ref{fig:TFG_pipeline} illustrates the standard talking face generation pipeline.

\noindent\textbf{Audio-Video Generation Setups.}
We define three complementary generation setups to probe different aspects of lip leakage.
In the first setup, silent-input generation (SI), videos are generated using silent audio.
The resulting lip movements are then evaluated with the original audio.
This setup isolates the effect of the identity reference on lip motion, revealing whether lip movements are influenced by the reference rather than audio.
In the second setup, Audio-Matched (AM), videos are generated using their corresponding GT audio. 
This serves as a baseline for standard lip-sync performance and visual quality.
The third setup is Audio-Mismatched (XM) scenario, where videos are generated using randomly paired, non-matching audio. 
Comparing lip-sync scores between the AM and XM setups allows us to determine whether the model follows the input audio or leaks information from the identity reference. 
High performance on AM but low performance on XM indicates difficulty in following the audio or suppressing lip leakage from the reference.

\noindent\textbf{Identity Reference Selection Strategies.}
The choice of identity reference can significantly affect lip leakage. 
We consider two strategies: Current Reference (CR) and Alternative Reference (AR).
In CR, the identity reference is the same as the masked input frame.
This scenario maximizes visual consistency between input and reference.
In AR, the identity reference is chosen according to method-specific strategies when available. Some approaches select multiple reference frames from the video, in which case we follow their setup. If no specific selection strategy is provided, the first frame is used as the reference.
This setup reflects typical deployment scenarios and allows evaluation under less constrained conditions.

\subsection{Lip leakage Assessment Metrics}
We evaluate lip leakage using four complementary metrics: Silent LSE-C, Silent LSE-D, Lip-Sync Discrepancy with CR, and Lip-Sync Discrepancy with AR.

\noindent\textbf{Silent LSE-C (LSE-C$_S$).}
We compute LSE-C~\cite{prajwal2020lip}, which is a lip-sync error confidence metric, on silent-input generations against the original audio using SyncNet features~\cite{chung2017out}.

\noindent\textbf{Silent LSE-D (LSE-D$_S$).}
We measure LSE-D~\cite{prajwal2020lip}, which evaluates lip-sync error distance, on silent-input generation with original audio of the employed videos as in LSE-C$_S$.
We obtain the multimodal features from SyncNet~\cite{chung2017out}.

\noindent\textbf{Lip-Sync Discrepancy with CR (LSD-CR).}
In this metric, 
we calculate the LSE-C and LSE-D metrics in AM and XM setups and compute the average distance as below:
\begin{equation}
    \textrm{LSD-CR} =  0.5 \times (|\textrm{C}^{CR}_{AM} - \textrm{C}^{CR}_{XM}| + |\textrm{D}^{CR}_{AM} - \textrm{D}^{CR}_{XM}|)
    \label{eq:lsd_cr}
\end{equation}
C and D refer to LSE-C and LSE-D metrics, respectively.
$CR$ indicates current reference strategy, while $AM$ and $XM$ denote Audio-Matched and Audio-Mismatched setups, respectively.
The LSD-CR metric quantifies lip leakage when the reference image matches the input. 
Higher values indicate greater lip leakage, with a minimum possible score of $0$.

\noindent\textbf{Lip-Sync Discrepancy with AR (LSD-AR).}
This metric is calculated similarly to LSD-CR, but using the AR reference selection setup instead of CR, capturing lip leakage under more typical reference choices. 
When a model specifies a reference selection strategy (e.g., random, multiple, or modified references), we follow it; otherwise, the first frame is used as the identity reference.
The formula is shown in Equation \ref{eq:lsd_ar}.
\begin{equation}
    \textrm{LSD-AR} =  0.5 \times (|\textrm{C}^{AR}_{AM} - \textrm{C}^{AR}_{XM}| + |\textrm{D}^{AR}_{AM} - \textrm{D}^{AR}_{XM}|)
    \label{eq:lsd_ar}
\end{equation}

\subsection{Visual Quality and Identity Preservation}
In addition to assessing lip leakage, our experimental setups allow a detailed analysis of visual quality and identity preservation under varying reference selection strategies. 
By evaluating standard metrics such as SSIM~\cite{wang2004image}, PSNR, FID~\cite{heusel2017gans}, and CSIM across different combinations of Audio-Matched (AM), Audio-Mismatched (XM), Current Reference (CR), Alternative Reference (AR), and Silent-Input (SI) generations, we can examine how models respond to reference variation. 
Our framework provides a unified approach to assess both lip-sync fidelity and reference-driven robustness, offering deeper insights into how models capture, transfer, and preserve facial identity across diverse conditions.

\begin{table}[]
    \centering
    \begin{tabular}{l|c}
    \toprule
    Method & Alternative Reference  \\
    \midrule
    Wav2Lip~\cite{prajwal2020lip}     & First frame \\
    TalkLip~\cite{wang2023seeing}     & First frame \\
    IPLAP~\cite{zhong2023identity}       & Multiple references \\
    AVTFG~\cite{yaman2024audio}       & First frame \\
    PLGAN~\cite{yaman2024audiodriventalkingfacegeneration}      & First frame \\
    Diff2Lip~\cite{mukhopadhyay2024diff2lip}    & First frame \\
    \bottomrule
    \end{tabular}
    \caption{We apply each model’s proposed reference selection method in the Alternative Reference scenario; if none is specified, the first frame is used as the reference.}
    \label{tab:alternative_reference}
\end{table}

\section{Experimental Results}
\label{sec:experiments}

\noindent\textbf{Dataset and Evaluation.}
We evaluate our methodology on the publicly available and commonly used talking face generation dataset, LRS2~\cite{LRS2}, following the standard pre-processing and provided train-test splits. 
For each method, we generate videos under the three audio–video setups: Silent-Input (SI), Audio-Matched (AM), and Audio-Mismatched (XM). 
Both Current Reference (CR) and Alternative Reference (AR) strategies are applied according to the method-specific guidelines.
If a model does not specify a reference frame selection method beyond using the current frame, we use the first frame as the identity reference in the AR setting.
Table \ref{tab:alternative_reference} summarizes the identity reference selection methods used under the AR setup.
For each generated video, we compute four lip leakage metrics, LSE-C$_S$, LSE-D$_S$, LSD-CR, and LSD-AR, alongside standard visual quality metrics (SSIM, PSNR, FID) and identity preservation (CSIM). 
For CSIM, features are extracted from generated and target faces using ArcFace~\cite{deng2019arcface}, and cosine similarity is computed. 
Lip-sync performance is evaluated using standard LSE-C and LSE-D~\cite{chung2017lip,prajwal2020lip} across all setups, and Mouth Landmark Distance (LMD)~\cite{chen2019hierarchical} is calculated in the AM scenario by computing the distance between generated and GT mouth landmarks.

\noindent\textbf{Silent-Input Generation (SI) Analysis.}
We present the visual quality, identity preservation, and lip-sync scores in Table \ref{tab:silent_audio_input}. 
Videos are generated using silent-input audio, and metrics are computed by comparing them with the original (GT) audio to evaluate whether the model preserves the original lip shapes or lip shape features. 
In the table, for each metric, the first score corresponds to the Alternative Reference (AR) setup, while the second score corresponds to the Current Reference (CR) setup.
From the results, except for Diff2Lip and TalkLip, most models exhibit similar visual quality performance in terms of SSIM and PSNR. 
In FID, TalkLip shows relatively larger changes. 
Regarding identity preservation (CSIM), TalkLip and Diff2Lip experience substantial performance degradation under the AR setup.
For lip-sync metrics, TalkLip and AVTFG achieve high confidence and low distance scores with the original audio, even though the audio was not provided during generation. 
This demonstrates severe lip leakage from the identity reference when using the CR setup. 
Using the AR setup can mitigate lip leakage; however, while AVTFG maintains robust performance, TalkLip suffers a significant drop in visual quality and identity preservation under AR conditions.

\begin{table}[]
    \centering
    \resizebox{\linewidth}{!}{\begin{tabular}{c|cccccccccccc}
    \toprule
        Method & \multicolumn{2}{c}{SSIM} & \multicolumn{2}{c}{PSNR} & \multicolumn{2}{c}{FID} & \multicolumn{2}{c}{LSE-C} & \multicolumn{2}{c}{LSE-D} & \multicolumn{2}{c}{CSIM} \\
        \midrule
        Wav2Lip    & 0.95 & 0.95 & 30.69 & 31.01 & 3.88 & 4.03 & 2.57 & 3.64 & 8.98 & 8.15 & 0.86 & 0.86 \\
        TalkLip    & 0.85 & 0.94 & 24.64 & 29.74 & 6.43 & 3.08 & 2.35 & 5.21 & 10.82 & 8.34 & 0.75 & 0.87 \\
        IPLAP      & 0.87 & 0.89 & 27.69 & 28.61 & 4.29 & 4.64 & 2.71 & 2.74 & 8.82 & 8.82 & 0.78 & 0.80 \\
        AVTFG      & 0.95 & 0.95 & 32.63 & 32.96 & 5.04 & 5.99 & 2.75 & 6.31 & 8.90 & 6.81 & 0.88 & 0.88 \\
        PLGAN & 0.94 & 0.95 & 31.27 & 31.59 & 3.74 & 5.07 & 2.70 & 2.93 & 9.02 & 8.51 & 0.86 & 0.87 \\
        Diff2Lip   & 0.86 & 0.93 & 26.09 & 30.52 & 3.36 & 3.37 & 2.95 & 2.79 & 10.21 & 9.52 & 0.76 & 0.84 \\
        \bottomrule
    \end{tabular}}
    \caption{Silent-input video generation results. Evaluation is done by employing original (GT) audio.}
    \label{tab:silent_audio_input}
\end{table}

\noindent\textbf{Audio-Mismatched (XM) Analysis.}
In Table \ref{tab:cross_test_set}, we present the evaluation results for the audio-mismatched (XM; cross-test). 
For each metric, the first column corresponds to the Alternative Reference (AR) setup, while the second column corresponds to the Current Reference (CR) setup.
Diff2Lip experiences a slight drop in lip-sync performance under the CR setup, whereas TalkLip shows a significant decrease. 
The strong performance of Diff2Lip under AR indicates that lip leakage primarily occurs when the model uses the input frame as the identity reference. 
When a different reference image is provided, Diff2Lip is able to rely less on identity-derived lip features and more on the driving audio.
In contrast, TalkLip exhibits poor lip-sync performance under both AR and CR conditions. 
IPLAP shows nearly identical performance between AR and CR; however, its LSE-C and LSE-D scores are extremely low, indicating that the model struggles to generate properly aligned lips even when the audio is provided.

\begin{table}[]
    \centering
    \resizebox{\linewidth}{!}{\begin{tabular}{c|cccccccccccc}
    \toprule
        Method & \multicolumn{2}{c}{SSIM} & \multicolumn{2}{c}{PSNR} & \multicolumn{2}{c}{FID} & \multicolumn{2}{c}{LSE-C} & \multicolumn{2}{c}{LSE-D} & \multicolumn{2}{c}{CSIM} \\
        \midrule
        Wav2Lip    & 0.84 & 0.84 & 24.62 & 25.84 & 3.39 & 7.89 & 7.98 & 7.35 & 6.79 & 7.18 & 0.74 & 0.83 \\
        TalkLip    & 0.85 & 0.93 & 25.70 & 29.11 & 4.04 & 2.89 & 6.04 & 4.80 & 8.21 & 9.40 & 0.74 & 0.86 \\
        IPLAP      & 0.86 & 0.89 & 28.99 & 29.85 & 3.95 & 3.98 & 3.63 & 3.71 & 10.10 & 10.02 & 0.77 & 0.80 \\
        AVTFG      & 0.83 & 0.85 & 24.18 & 26.43 & 5.32 & 5.78 & 6.90 & 6.84 & 8.63 & 7.90 & 0.72 & 0.72 \\
        PLGAN      & 0.86 & 0.89 & 25.38 & 27.66 & 4.99 & 4.11 & 7.95 & 7.58 & 6.64 & 6.81 & 0.73 & 0.73 \\
        Diff2Lip   & 0.86 & 0.92 & 25.49 & 30.32 & 2.49 & 3.59 & 7.62 & 6.71 & 6.59 & 7.26 & 0.76 & 0.83 \\
        \bottomrule
    \end{tabular}}
    \caption{Quantitative results on Audio-Mismatched (XM) setup.}
    \label{tab:cross_test_set}
\end{table}

\noindent\textbf{Audio-Matched (AM) Analysis.}
In this setup, we follow the most common evaluation protocol in the talking face generation literature and report the results in Table \ref{tab:test_set}. 
Videos are generated using the GT audio-video pairs. 
While this evaluation is standard, it can be misleading when models exhibit lip leakage. 
In such cases, models may achieve very high performance under the CR setup, but their performance drops noticeably under the AR setup. 
This highlights the importance of also evaluating models under the XM setup to obtain a more robust assessment. 
In our experiments, TalkLip and Diff2Lip show a significant decrease in both visual quality and identity preservation metrics under AR and XM conditions, revealing vulnerabilities that are not captured by the standard evaluation alone.

\begin{table}[]
    \centering
    \resizebox{\linewidth}{!}{\begin{tabular}{c|cccccccccccccc}
    \toprule
        Method & \multicolumn{2}{c}{SSIM} & \multicolumn{2}{c}{PSNR} & \multicolumn{2}{c}{FID} & \multicolumn{2}{c}{LMD} & \multicolumn{2}{c}{LSE-C} & \multicolumn{2}{c}{LSE-D} & \multicolumn{2}{c}{CSIM} \\
        \midrule
        Wav2Lip & 0.86 & 0.95 & 26.53 & 31.01 & 7.05 & 3.97 & 2.38 & 1.15 & 7.59 & 7.73 & 6.75 & 6.44 & 0.84 & 0.86 \\
        TalkLip & 0.86 & 0.94 & 26.11 & 29.89 & 4.94 & 2.99 & 2.34 & 1.28 & 8.53 & 9.27 & 6.08 & 5.54 & 0.75 & 0.87 \\
        IPLAP   & 0.88 & 0.87 & 27.99 & 29.67 & 3.78 & 4.10 & 2.34 & 2.11 & 5.96 & 6.49 & 7.54 & 7.16 & 0.79 & 0.82 \\
        AVTFG   & 0.95 & 0.95 & 32.63 & 31.27 & 5.06 & 4.51 & 1.13 & 1.19 & 7.94 & 7.95 & 6.35 & 6.30 & 0.88 & 0.80 \\
        PLGAN   & 0.94 & 0.95 & 31.27 & 32.64 & 4.62 & 3.83 & 1.16 & 1.13 & 7.68 & 8.41 & 6.43 & 6.03 & 0.86 & 0.79 \\
        Diff2Lip& 0.87 & 0.94 & 26.12 & 31.68 & 2.63 & 3.80 & 2.12 & 1.50 & 7.82 & 7.87 & 6.48 & 6.46 & 0.78 & 0.85 \\
        \bottomrule
    \end{tabular}}
    \caption{Evaluation results on Audio-Matched (AM) setup.}
    \label{tab:test_set}
\end{table}

\begin{table}[]
    \centering
    \footnotesize
    \begin{tabular}{c|cccc}
    \toprule
        Method & LSE-C$_s$ $\downarrow$ & LSE-D$_s$ $\uparrow$ & LSD-CR $\downarrow$ & LSD-AR $\downarrow$ \\ 
        \midrule
        Wav2Lip    & 3.64 & 8.15 & \textbf{0.56} & 0.22 \\
        TalkLip    & 5.21 & 8.34 & 4.16 & 2.31 \\
        IPLAP      & \textbf{2.74} & 8.82 & 2.82 & 2.45 \\
        AVTFG      & 6.31 & 6.81 & 1.36 & 1.66 \\
        PLGAN      & 2.93 & 8.51 & 0.80 & 0.24 \\ 
        Diff2Lip   & 2.79 & \textbf{9.52} & 0.98 & \textbf{0.15} \\ 
        \bottomrule
    \end{tabular}
    \caption{Evaluation results with the lip leakage metrics.}
    \label{tab:leaking}
\end{table}

\noindent\textbf{Lip Leakage Metrics.}
In Table \ref{tab:leaking}, we report the scores of each model using our proposed lip-leakage assessment metrics. 
According to the results, TalkLip and AVTFG exhibit the poorest performance, whereas PLGAN, Diff2Lip, and Wav2Lip achieve more accurate results. 
It is important to note that LSE-C$_S$ and LSE-D$_S$ metrics do not reflect the models’ overall performance. 
For instance, IPLAP achieves high performance according to these metrics, as it does not exhibit lip leakage under the silent-input condition. 
However, when evaluated with LSD-CR and LSD-AR, IPLAP shows a significant performance drop, revealing clear identity-driven lip leakage. 
These observations demonstrate that all proposed metrics provide complementary insights, collectively offering a more comprehensive evaluation of lip leakage.

\noindent\textbf{Identity Reference Selection.}
Based on our detailed experiments and analysis, we found that the most effective identity reference selection method for maximizing visual quality and stability while minimizing lip leakage is to use multiple reference images with different poses~\cite{zhong2023identity}. 
Furthermore, selecting a reference image whose lip appearance is most dissimilar~\cite{zhang2024musetalk} from the GT during training, or using a silent-face~\cite{yaman2024audiodriventalkingfacegeneration,yaman2025mask} or stabilized-face image~\cite{cheng2022videoretalking} as the identity reference, are effective strategies for reducing lip leakage.

\section{Conclusion}
We presented a systematic framework to analyze lip leakage from identity reference image for talking face generation. 
Our methodology combines complementary generation setups (Silent-Input, Audio-Matched, and Audio-Mismatched), reference selection strategies (Current vs. Alternative), and tailored evaluation metrics. 
In addition to adapting LSE-C and LSE-D to reveal leakage, we introduced Lip-Sync Discrepancy (LSD) score. 
Beyond lip-sync evaluation, our setups also enable analysis of visual quality and identity preservation under different reference strategies, offering deeper insights into model robustness. 
Together, these contributions establish a model-agnostic assessment protocol that uncovers subtle but important weaknesses overlooked by conventional metrics, and provide a stronger benchmark in talking face generation.
As future work, additional recent methods such as LatentSync~\cite{li2024latentsync}, MuseTalk~\cite{zhang2024musetalk}, and OmniSync~\cite{peng2025omnisync} can be evaluated.

\noindent\textbf{Acknowledgment.}
This work was supported in part by the EU’s Horizon research and innovation program, project: Meetween (101135798). Development of video face dubbing tools made under contract from KIT Campus Transfer GmbH (KCT).

%
%
%



\bibliographystyle{IEEEbib}
\bibliography{strings,refs}

\end{document}